\documentclass[runningheads]{llncs}

\usepackage{eccv}

\usepackage{eccvabbrv}
\usepackage{graphicx}
\usepackage{booktabs}

\usepackage{algorithm}
\usepackage{algpseudocode}
\usepackage{multirow, booktabs}
\usepackage{siunitx}
\usepackage{multirow}
\usepackage{tabularx}
\usepackage{microtype}
\usepackage{mathtools}
\usepackage{xargs}     
\usepackage{float}
\usepackage{makecell}
\usepackage{blindtext}% for dummy text only
\usepackage{graphicx}
\usepackage{xcolor}

\usepackage[accsupp]{axessibility}  % Improves PDF readability for those with disabilities.

\definecolor{DarkGreen}{rgb}{0.1,0.6,0.1}
\def\check c{{\color{DarkGreen}\checkmark}}
\def\x x{{\color{red}x}}
\newcommand{\AlgName}{STREAM}
\newcommand{\DataName}{Motorica++}

\usepackage[breaklinks,colorlinks,citecolor=eccvblue]{hyperref}
\usepackage{orcidlink}

\begin{document}

\title{Text Dictates, Music Decorates: Energy-based Attention for Editable Dance Motion Generation} 

\titlerunning{STREAM}

\author{Seong Jong Yoo \and
Siyuan Peng \and
Felix Gu \and
Stratis Aloimonos \and
Cornelia Ferm\"{u}ller}

\authorrunning{S.J. Yoo, et al.}

\institute{University of Maryland, College Park, MD, 20742, USA \\
Perception and Robotics Group %\\ 
% \email{$\{\text{yoosj, peng2000} \}$@umd.edu, felixgu12@gmail.com,}
}

\maketitle

\begin{abstract}
Choreographic motion generation poses unique challenges for AI, demanding precise semantic control over complex, temporally structured, and expressive full-body dynamics. While existing models can synthesize motion from music, they remain largely black boxes. Conversely, attempting to condition generation on both text and music frequently leads to modality collapse, where dense acoustic rhythms overwhelm sparse semantic text prompts, destroying user controllability. To resolve this spatial-temporal conflict, we propose STREAM (Structural-Temporal Rhythmic Energy-based Attention for Motion), a modality-decoupled diffusion transformer. STREAM strictly separates conditioning pathways: global text semantics dictate the kinematic structure via Adaptive Layer Normalization (AdaLN), while a novel Bimodal Energy-Based Attention Module (BEAM) routes these features to the musical beat without overwriting the semantics. We further introduce Motorica++, a newly curated dataset enriched with domain-specific dance vocabulary and frame-level semantic annotations from existing Motorica dataset. Additionally, to rigorously quantify zero-shot editability, we propose the Exchange Evaluation Protocol and Editable Dance Score (EDS). Through extensive experiments, STREAM achieves state-of-the-art alignment between motion and music while preserving choreographic semantics, positioning AI not merely as a reactive synthesizer, but as a controllable, collaborative partner for artistic direction. The source code and datasets are available at https://github.com/SeongJong-Yoo/STREAM.
\end{abstract}
\section{Introduction}
\label{sec:intro}
Dance is a universal phenomenon across cultures, ethnicities, and eras~\cite{sieversMusicMovementShare2013, bassoDanceBrainEnhancing2021, cameronCrossculturalInfluencesRhythm2015}. Humans possess a strong innate desire to move their bodies with music and rhythm~\cite{kaepplerDanceEthnologyAnthropology2000}. 
When these spontaneous movements are shaped into structured, intentional forms, they become artistic expressions crafted through choreographic design~\cite{angelovYouChoreographerCreating2023}. Professional choreographers translate musical ideas into movement, balancing artistic vision with the physical realities of the human body.

Although AI-driven dance motion generation has achieved significant advances in realism and synchronization, current systems still lack one crucial capability: semantically meaningful control~\cite{lodge, aistpp, LDA, peng2024choreographing}. Most generative models operate as black boxes dominated by given music conditions, offering limited access to the nuanced, interpretable directions that choreographers rely on. 
Even existing controllable (editable) approaches typically fall into one of three categories: micro-level edits~\cite{tseng2023edge} (\eg, joint-wise adjustments, which are tedious and counter-intuitive), temporal inpainting ~\cite{huang2024beat, tseng2023edge} (which cannot specify semantic intent), and high-level conditioning~\cite{LDA} (\eg, genre labels that offer minimal fine-grained control). These constraints make current systems misaligned with creative workflows in which choreographers require precise, expressive, and concept-driven manipulation of movement. 

\begin{figure}[t]
	\centering
	\includegraphics[width=0.95\linewidth]{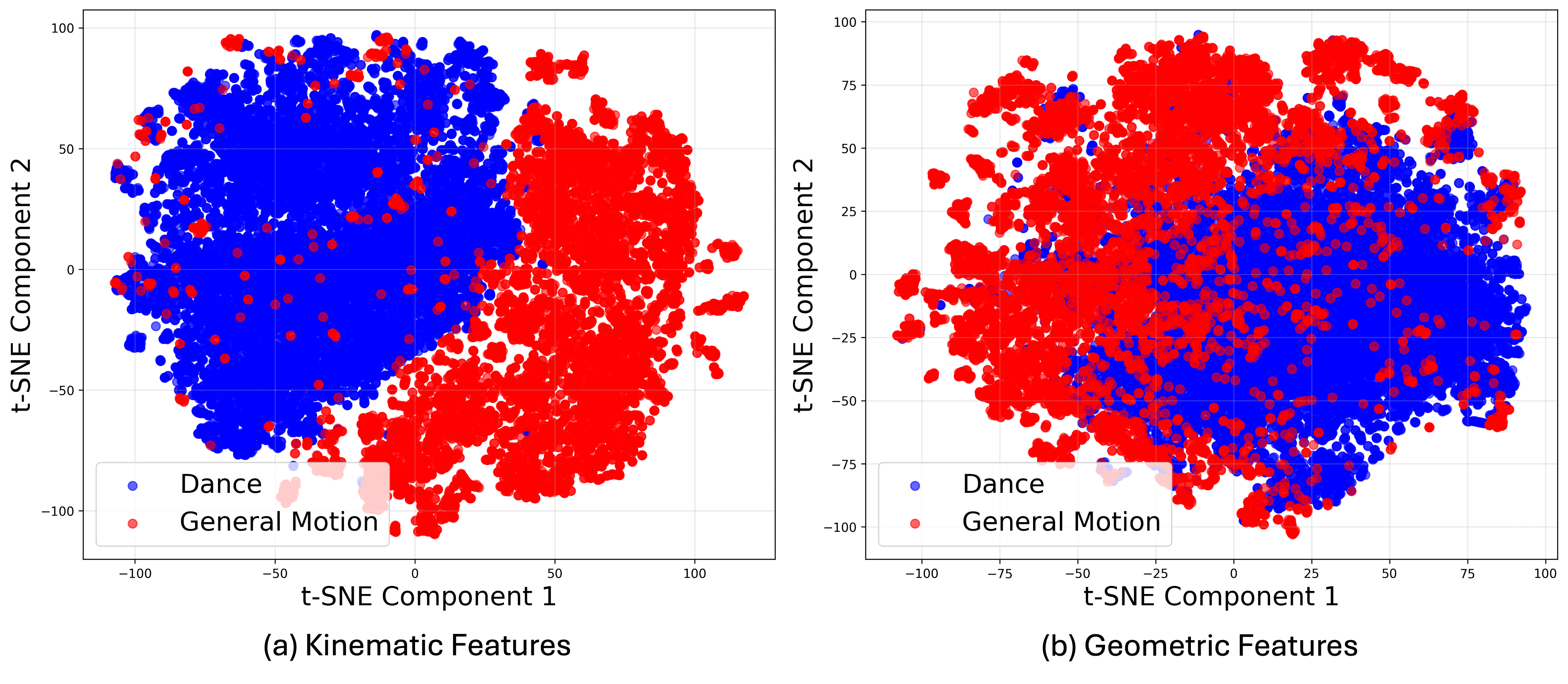}
	\caption{
	     t-SNE visualization of dance (Motorica) and general motion (HumanML3D) of kinematic features~\cite{kinetic_feature} and geometric features~\cite{geometric_feature}. 
	}
	\label{fig:features}
\end{figure}

In contrast to dance-specific generative models, controllability in text-to-motion generation has advanced rapidly~\cite{tevetHumanMotionDiffusion2022a, athanasiouMotionFixTextDriven3D2024, goelIterativeMotionEditing2024, Huang_2024_ECCV, EnergyMoGen}. These models achieve fine-grained control through textual descriptions, enabled by large-scale text-motion datasets such as HumanML3D~\cite{HumanML3D}, KIT-ML~\cite{KIT}, and Motion-X~\cite{linMotionXLargescale3D2023a}. 
Recent efforts further attempt to unify multiple conditioning modalities~\cite{liGENMOGENeralistModel2025} or combine general text-motion datasets with dance datasets to increase controllability~\cite{TM2D, yangUniMuMoUnifiedText2025, DanceEditor}. However, the dynamics of everyday human motions differ fundamentally from the structure, expressiveness, and rhythmic constraints of dance, as shown in Fig.~\ref{fig:features}. When standard cross-attention architectures attempt to fuse these modalities, they often suffer from \textbf{Modality Collapse}, in which the dense, high-frequency rhythmic signals of the music overwhelm the sparse, high-level semantic signals of the text, causing the network to ignore the user's prompt and revert to music-conditioned dance generation. 

To bridge this gap, we propose a system built on three components: a professionally curated dataset, a novel energy-based architecture for controllable dance generation that is directed by text and modulated by music, and a new metric to quantify controllability in dance generation task. The system is designed to speak the language of choreographers by providing a direct, semantically meaningful interface while preserving high fidelity and strong alignment with musical structure. By making dance generation controllable, learnable, and artist-friendly, we aim to empower creators rather than replace them. 

Specifically, we first {\bf annotate the Motorica dance dataset~\cite{LDA}} with frame-level dance-technique labels curated by a professional dancer and detailed human-motion text descriptions. While datasets such as AIST++~\cite{aistpp}, Motorica~\cite{LDA}, FineDance~\cite{liFineDanceFinegrainedChoreography2023}, and DanceRemix~\cite{DanceEditor} contain high-quality music-motion pairs, they lack fine-grained text-motion annotations. Our \DataName \ dataset addresses this deficiency by linking dance motion with dance-specific textual descriptions. 
Second, we propose \textbf{\AlgName}, a Structural-Temporal Rhythmic Energy-based Attention Module for the dance Motion generation pipeline. To prevent modality collapse, \AlgName \ strictly disentangles the conditioning pathways, \textit{text dictates}, while \textit{music decorates}. Specifically, we inject global text semantics via Adaptive Layer Normalization (AdaLN) to define the spatial kinematic manifold, while our novel \textbf{Bimodal Energy-based Attention Module (BEAM)} injects raw acoustic energy directly into the attention logits. 
By construction, the music pathway conveys only the temporal self-similarity structure of the audio: the alignment drift term depends on the music solely through $S_m$, which encodes when the motion should repeat or vary, but carries no semantic content that could override the text conditioning. 
% This design encourages the motion to follow the musical rhythm while preserving the user's semantic commands, which we validate empirically in ~\cref{sec:abl}.
Finally, we propose the  {\bf Exchange Evaluation Protocol} and a unified metric, the {\bf Editable Dance Score (EDS)}. Existing editable dance generation pipelines rarely evaluate zero-shot editability under conflicting conditions (\eg, forcing a slow semantic dance onto a fast acoustic beat). By evaluating models on mismatched text-music pairs, EDS computes the harmonic mean of semantic preservation and rhythmic adaptation, rigorously penalizing models that suffer from modality collapse. In summary, our contributions are as follows:
\begin{enumerate}
    \item We curate a domain-specific dance dataset, labeled with frame-level dance-technique annotations and detailed motion text descriptions collected by a professional dancer, addressing the lack of fine-grained text-motion annotations in existing dance datasets. 
    \item We propose \AlgName, a text-controlled, music-decorated pipeline for generating dance motions, designed with a novel Bimodal Energy-based Attention Module. \AlgName \ enables semantic control of dance motion through text while decorating it to follow musical structure and beats.
    \item We introduce a new metric, called Editable Dance Score (EDS). This experiment and metric are designed to measure a new task for semantically controlled yet musically aligned motion generation.
\end{enumerate}
\section{Related Works}
\label{sec:related_works}
\subsection{Dance Motion Generation}
The key challenges include ensuring physical plausibility, producing diverse and expressive motions, achieving fine-grained controllability, and modeling complex human-body interactions. 
To address them, prior work has explored conditioning modalities such as music~\cite{aistpp, sunYouNeverStop2022}, style (genre)~\cite{LDA, lodge}, and text~\cite{TM2D, yangUniMuMoUnifiedText2025}. 
\textbf{Music-driven dance generation} focuses on synthesizing motion coherent with musical structure. For example, EDGE~\cite{tseng2023edge} uses a transformer-based diffusion model conditioned on music features extracted from Jukebox~\cite{dhariwalJukeboxGenerativeModel2020}, and \cite{shah2025dancemosaic} developed DanceMosaic, which incorporates music in its masked motion modeling. 
\textbf{Style-conditioned dance generation} aims to generate motion aligned with specific dance styles (\eg, ballet, hip-hop). \cite{wang2024flexible} proposed DGSDP, a diffusion model guided by textual style prompts. LDA~\cite{LDA} conditioned on both music and style, applied classifier-free guidance~\cite{hoClassifierFreeDiffusionGuidance2022} to enable smooth interpolation between different styles.
\textbf{Language-Based Dance Motion Generation} leverages detailed motion descriptions from general text-motion datasets, such as HumanML3D~\cite{HumanML3D}. 
For instance, TM2D \cite{gong2023tm2d} presented a novel approach to generating 3D dance movements using a VQ-VAE conditioned on text and music modalities. DanceEditor created dance pairs with difference descriptions, allowing text prompted dance motion editing~\cite{DanceEditor}. 
% \citet{swdance} utilized a transfer-learned diffusion model to demonstrate the potential of aiding dance practice using spoken word text. 

\begin{figure*}[t]
	\centering
	\includegraphics[width=\linewidth]{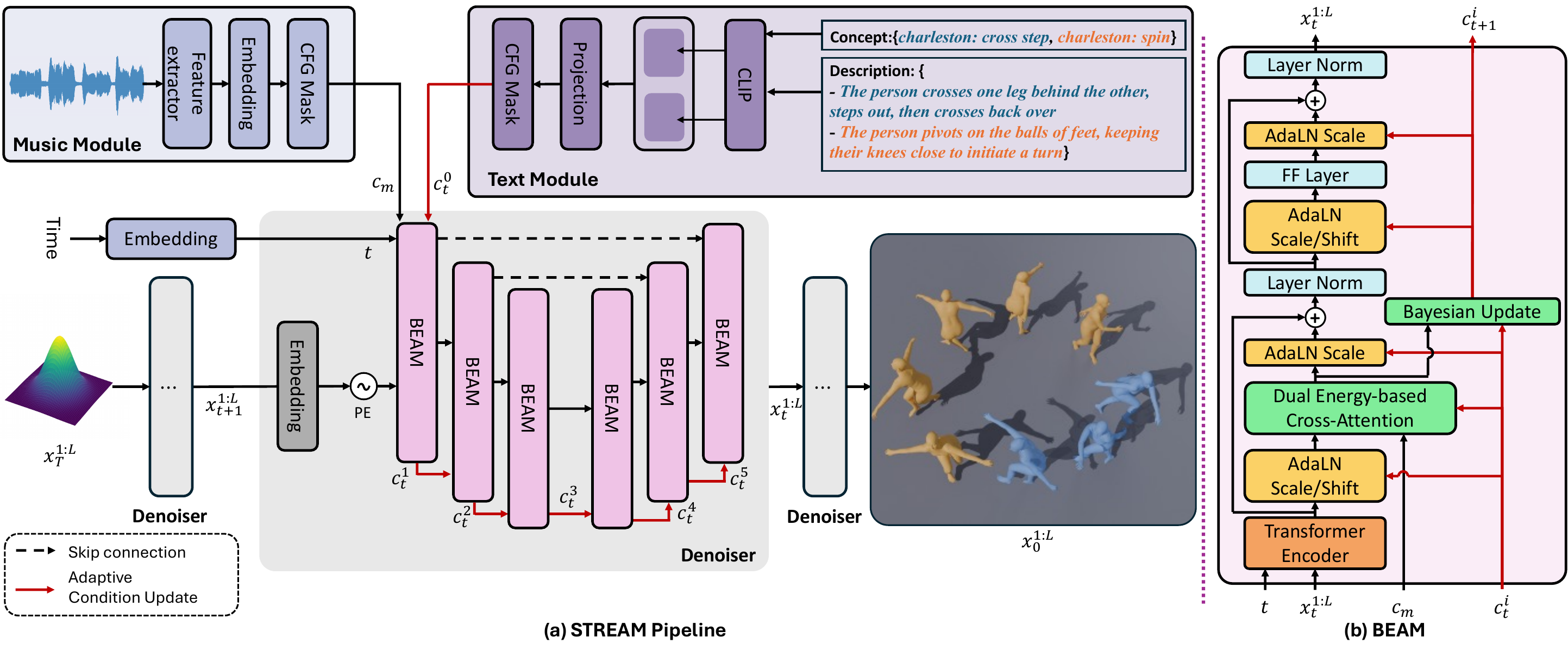}
	\caption{
	    Overview of \AlgName. (a) \textbf{Left}: \AlgName \ is first controlled by text (high-level concept and a low-level detailed description). Second, the music condition modulates the motion via the music alignment energy function, transforming general motion into \textit{dance-like} motion aligned to musical beats. (b) \textbf{Right}: The Bimodal Energy Attention Module (BEAM) adaptively updates the text conditions (red line) via MAP estimation, and global text information is applied through AdaLN modulation. 
	}
	\label{fig:pipeline}
\end{figure*}

Despite these advances, existing methods still struggle with limitations such as insufficient fine-grained control \cite{shah2025dancemosaic, peng2024choreographing}, and a persistent semantic gap between textual descriptions and the nuanced expressiveness of dance~\cite{TM2D,yangUniMuMoUnifiedText2025,attributeGeneration}. Our method bridges this gap by introducing a controllable, text-guided framework that captures fine-level dance semantics while maintaining strong alignment with music. Also, \AlgName\ enables interpretable motion control, preserves dance structure, and produces expressive movements that better reflect choreographic intent.

\section{Preliminaries}

\textbf{Energy-Based Models (EBMs)} define a probability distribution over data via a scalar energy function, expressing the density in the exponential family form:

\begin{equation}
p_\theta(x) = \frac{1}{Z(\theta)} \exp(-E_\theta(x)),
\end{equation}
where $\theta$ denotes model parameters, $E_\theta(x): \mathbb{R}^d \to \mathbb{R}$ is the scalar energy function, and
$Z(\theta)=\int \exp(-E_\theta(x)) dx$ is the partition function.
Training aims to shape $E_\theta(x)$ so that data samples obtain low energy while unobserved configurations map to high energy.
Learning an explicit energy landscape also enables compositional reasoning, as distinct concepts can be combined through Boolean-like operations~\cite{duCompositionalVisualGeneration2020, hintonTrainingProductsExperts2002, EnergyMoGen}. Furthermore, this formulation allows for inference-time optimization to drive iterative refinement~\cite{duLearningIterativeReasoning2024, gladstoneEnergyBasedTransformersAre2025}.
Finally, because EBMs impose no structural assumptions on the form of $E_\theta$, they offer strong modeling flexibility~\cite{songHowTrainYour2021}.

A major challenge is the intractable partition function, which complicates sampling and training~\cite{duReduceReuseRecycle2024}.
To mitigate this, several approximations are used:
(1) Contrastive Divergence (CD) employs MCMC sampling to estimate likelihood gradients~\cite{nealMCMCUsingHamiltonian2011, hintonTrainingProductsExperts2002};
(2) Score Matching (SM) learns the score function $s_\theta(x)=\nabla_x\log p_\theta(x)=-\nabla_xE_\theta(x)$, avoiding the partition term~\cite{hyvarinenEstimationNonnormalizedStatistical2005, songScoreBasedGenerativeModeling2021}; and
(3) Noise Contrastive Estimation (NCE) reformulates learning as classifying data versus noise samples~\cite{gutmannNoisecontrastiveEstimationNew2010, duLearningIterativeReasoning2024}.

\noindent \textbf{Relation to Hopfield Networks and Self-Attention:}
\cite{ramsauerHopfieldNetworksAll2021} showed that self-attention can be viewed as minimizing an energy function of a modern Hopfield network:

\begin{equation}
E(X, \xi) = -\text{lse}(\beta X^\top\xi) + \tfrac{1}{2}\xi^\top\xi + C,
\label{eq: modern_energy}
\end{equation}
where $\text{lse}(X)=\log\sum_i \exp(x_i)$ and $\xi\in\mathbb{R}^{M\times d}$ is the state.
Minimizing $E$ using the Concave–Convex Procedure (CCCP)~\cite{yuilleConcaveConvexProcedureCCCP2001} yields the update
\begin{equation}
\xi_{\text{new}} = X\text{Softmax}(\beta X^\top \xi),
\label{eq: update}
\end{equation}
which corresponds exactly to the self-attention mechanism, with $X$ as key/value, $\xi$ as query, and $\beta = 1/\sqrt{d}$.
This establishes a direct theoretical link between EBMs and attention-based architectures. Adapting this perspective, we propose a new energy functions with both text and music conditions. By minimizing the new energy function, \AlgName \ is able to have semantic control while preserving musical alignment.

\section{Method}
In this section, we present the pipeline for the diffusion-based Structural-Temporal Energy-based Attention for dance Motion (\AlgName) generation pipeline (Fig.~\ref{fig:pipeline}). 
\AlgName\ strictly enforces the spatial and semantic structure of the human motion (the "What") as specified by the text descriptions. The model explicitly aligns the generated motion with the acoustic cues (the "When"), ensuring that the music modulates the temporal dynamics of the semantic motion. By decoupling the structural semantics from the rhythmic timing via the Bimodal Energy-based Attention Module (BEAM), \AlgName \ prevents modality collapse and enables semantically controllable, yet musically aligned, motion generation.

% enables controllable motion synthesis from both high-level choreographic concepts and fine-grained motion descriptions, while enforcing dance-like dynamics aligned with the musical beat through Bimodal Energy-based Attention Module (BEAM). The forward path of BEAM lowers the energy functions, defined by semantic (text) and musical alignment, instantiated by Modern Hopfield Networks.

\subsection{Problem Statement}
Our goal is to generate a human dance motion $x^{1:L}$ of length $L$, conditioned on music $c_{m}$ and text $c_{t}=\{c_h, c_l\}$, where $c_h$ represents a high-level concept such as style, dance technique, or action label, and $c_l$ is a low-level detailed description of human motion. 
The two modalities play distinct roles: the text conditions the motion behavior, while the music modulates temporal rhythm and beat alignment. We model the conditional distribution  $p(x^{1:L}|c_{t}, c_{m})$ using a denoising diffusion probabilistic model~\cite{ddpm} formulated from an energy-based perspective~\cite{ramsauerHopfieldNetworksAll2021, parkEnergyBasedCrossAttention2023}, which allows the system to capture structured dependencies between motion, text, and music cues.

\subsubsection{Motion Representation:}
We represent a motion sequence of length $L$ as $x^{1:L}=\{x^1 ,\cdots, x^L \}\in\mathbb{R}^{L\times D}$, sampled at 30 FPS over 5 second clips. Each frame $x^i$ follows the SMPL parameterization~\cite{SMPL}, defined as $x^i=(t_g^i,\alpha^i,\beta^i)$, where $\alpha^i\in\mathbb{R}^{24\times6}$ denotes joint rotations in 6D representation~\cite{zhouContinuityRotationRepresentations2019}, $t_g^i\in\mathbb{R}^3$ is the global root translation, and $\beta^i\in\mathbb{R}^{10}$ encodes the body shape.
Following prior works~\cite{NIFTY, leeMultiActLongTerm3D2023, tripathiHUMOSHumanMotion2024, athanasiouMotionFixTextDriven3D2024}, all motions are canonicalized with respect to the first frame $x^1$, oriented to face the forward direction, and aligned such that the floor lies on the positive $xy$-plane. 

\subsection{\AlgName \ Pipeline}
\subsubsection{Diffusion Framework:}
Following the great success of diffusion frameworks for human motion generation tasks~\cite{chenExecutingYourCommands2023, tevetHumanMotionDiffusion2022a, EnergyMoGen, LDA, tseng2023edge}, \AlgName \ uses the DDPM~\cite{ddpm} framework to model $p(x|c_{t},c_{m})$.
The forward diffusion process progressively adds noise to a data sample, $x_0\sim p(x)$, according to a predefined noise scheduler, $\alpha_t$, yielding $x_T\sim\mathcal{N}(0, \textbf{I})$ at time $T$: 
\begin{equation}
    p(x_t|x_{t-1})=\mathcal{N}(x_t; \sqrt{\alpha_t}x_{t-1}, (1-\alpha_t)\textbf{I})
\end{equation}
Then the reverse process, parameterized by $\theta$, denoises samples  step-by-step using a Gaussian transition~\cite{diffuison2015, ddpm}:
\begin{equation}
    p_{\theta}(x_{t-1}|x_t)= \mathcal{N}(x_{t-1};\mu_\theta(x_t, t) ,\Sigma_\theta(x_t, t) \textbf{I})
\end{equation}
Our denoiser adopts a UNet-like architecture~\cite{UNET, chenExecutingYourCommands2023} augmented with multiple Bimodal Energy-based Attention Module (BEAM) layers. 
The BEAM consists of three components: (1) a Text Adaptive Layer Normalization (Text-AdaLN), (2) a Dual Energy-based Cross-Attention (D-EBCA), and (3) a Bayesian Update Module. 
Text-AdaLN modulates global text-condition information into motions, and then the D-EBCA minimizes the energy function defined by the text, music, and motions. Finally, the abstract text condition is optimized via MAP estimation~\cite{EnergyMoGen, parkEnergyBasedCrossAttention2023} as shown in ~\cref{fig:pipeline}. (b).
% Unlike ~\cite{EnergyMoGen}, our BEAM effectively integrates both music and text conditions via the Deep Transformer, enabling the text to control motion behavior while the music enriches its expressive details.

We further employ classifier-free guidance (CFG)~\cite{hoClassifierFreeDiffusionGuidance2022} for multi-conditioned motion generation. During training, we randomly mask the text condition and music condition with probabilities $p_{cfg_t}, p_{cfg_m}$, respectively.
To ensure the music condition \textit{modulates} the rhythm without overpowering the semantic structure during inference, we employ Hierarchical Classifier-Free Guidance, by guiding text semantic with rhythmic delta. In this way, we guarantee that the semantic manifold is established before rhythmic modulation is applied. Specifically, our inference time CFG equation is:
\begin{align}
    \hat{x}_{t-1} = \hat{x}_{\theta}(x_{t}) 
    +\lambda_t(\hat{x}_{\theta}(x_t, c_{t}) - \hat{x}_{\theta}(x_t))
    +\lambda_m(\hat{x}_{\theta}(x_t,  c_{t}, c_m ) - \hat{x}_{\theta}(x_t, c_t ))
    \label{eq:cfg}
\end{align}
where $\lambda_t$ and $\lambda_m$ are guidance hyperparameters of text and music, respectively with condition of $\lambda_t > \lambda_m$.

\subsubsection{Condition Modules:}
As outlined in the previous section, the music and text conditions serve different purposes: the text controls semantic generation, and the music refines the motion's temporal and stylistic details, transforming it into a dance. 
% To capture this distinction, \AlgName~disentangles two conditions via separate conditioning pathways and a motion neutralization module.
The music module extracts features from each music clip using a pretrained Jukebox model~\cite{dhariwalJukeboxGenerativeModel2020} sampled at 30 Hz~\cite{tseng2023edge, luoPOPDGPopular3D2024}. The extracted features are projected to form the music condition embedding $c_{m}\in\mathbb{R}^{L\times D}$. 
On the other hand, the text condition comprises both high-level information and low-level descriptions, denoted by $c_{t}=\{c_h, c_l\}$. We encode $c_h$ and $c_l$ using a pretrained CLIP text encoder~\cite{CLIP}, concatenate to project them onto an initial embedding  $c_{t}^0\in\mathbb{R}^{L\times D}$, as shown in ~\cref{fig:pipeline}. 
This embedding is iteratively refined through the BEAM layers, enabling hierarchical alignment between linguistic cues and motion features.
Consequently, $c_t$ effectively captures both global motion semantics (\eg, "Charleston cross step") and detailed movement description (\eg, "The person crosses one leg behind the other" shown with blue letters in ~\cref{fig:pipeline}).

\subsection{Bimodal Energy-based Attention Module (BEAM)}
% The main goal of BEAM is how to properly disentangle text and music conditions while infusing their property to motion generation task. 
Text contains abstract and high-level information. In choreography, body movements are highly dense and complex, unlike everyday human motion, \eg, HumanML3D~\cite{HumanML3D}. Dancers do not move randomly; their movements follow well-defined  structural conventions, commonly referred to as dance technique. While music provides dynamic, fine-grained local cues, the dancers respond to instantaneously, it often carries stronger correlational signals than text in dance motion generation tasks. As a result, models tend to ignore the textual condition and rely predominantly on musical cues. To address this imbalance, we propose the Bimodal Energy-Based Attention Module, which structurally enforces the disentanglement of text and music conditions through an inductive bias.

\subsubsection{Text-AdaLN:}
Our text condition comprises both high-level information and low-level descriptions of body movements. Therefore, we apply text conditions globally to motion query through AdaLN~\cite{AdaLN}. Specifically, the motion query $x$ is modulated as $x'=\gamma(c_t)\odot \text{LayerNorm}(x)+\beta(c_t)$, where the scale $\gamma$ and shift $\beta$ are linearly projected from the text condition. In this way, the global text information physically transforms the query latent vector onto the correct semantic manifold. We apply text-AdaLn twice: once after the Transformer encoder with previous text conditions $c_t^i$, and the other after Dual Energy-based Cross-Attention with newly updated text conditions $c_t^{i+1}$. 

\begin{figure}[t]
	\centering
	\includegraphics[width=0.95\linewidth]{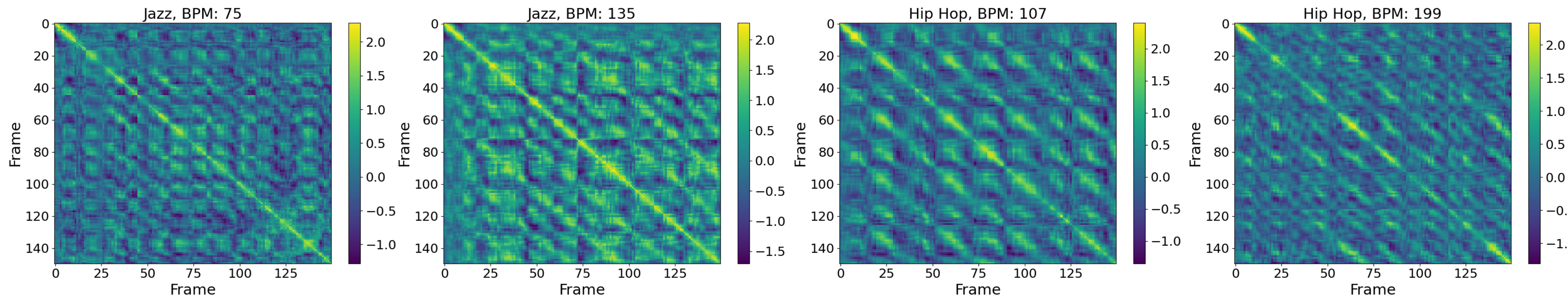}
	\caption{
	     Visualization example of self-similarity matrices of different music and BPM at first layer of D-EBCA. 
	}
	\label{fig:ssm}
\end{figure}

\subsubsection{Dual Energy-Based Cross Attention:}
\label{sec:DEBCA}
The Dual Energy-Based Cross Attention (D-EBCA) forms the core of \AlgName, coupling motion dynamics with textual semantics and aligning them musically through an energy-minimization formulation. 
Inspired by~\cite{parkEnergyBasedCrossAttention2023}, we reinterpret cross-attention as an \textit{energy-based model (EBM)}~\cite{ramsauerHopfieldNetworksAll2021}, where attention weights are optimized to minimize an energy function that encodes both \textit{semantic alignment} and \textit{structural regularity}. 
% This formulation enables iterative refinement of text embeddings via MAP updates, leading to semantically consistent and interpretable motion–text alignment.
Specifically, query, key, and value are defined as
\[
Q = xW_{Q}, \quad K_t = c_tW_{K^t}, \quad K_m=c_mW_{K^m}, \quad V_t = c_tW_{V},
\]
where $x$ denotes motion features, $c_t$ is the previous text embedding, and $c_m$ is music embedding.
We define the energy functions as:

\begin{align}
E(Q;K_t,K_m) &= - \frac{1}{\beta}\sum_{l=1}^L \text{lse}(A_l, \beta) + \mathcal{R}(Q, K_m) \nonumber \\ 
& +\frac{\alpha_t}{2}\|K_t\|^2 +\frac{\alpha_m}{2}\|K_m\|^2 +\frac{1}{2}\|Q\|^2
    \label{eq:dual_energy}
\end{align}

\begin{align}
    \mathcal{R}(Q, K_m)&=-\sum_{i,j}(S_m)_{i,j}\cdot(q_i^\intercal q_j)
    \label{eq:music_energy}
\end{align}
where $A=QK_t^\intercal+\gamma_m QK_m^\intercal$, $\text{lse}(\cdot)$ denotes  the log-sum-exp operator, and $S_m$ is the self-similarity matrix~\cite{SSM} of $K_m$. 
$\mathcal{R}(Q, K_m)$ is the music alignment energy function, measuring how well musical information is aligned with the given motion query.
Minimizing ~\cref{eq:dual_energy} yields the forward path:
\begin{align}
    Q_{new}=Q - \eta\nabla_Q E(Q, K_t, K_m)\approx\underbrace{\text{Softmax}(d^{-1/2}A) V_t}_{\text{Dual Attractor}}- \underbrace{\gamma_d\nabla_Q\mathcal{R}(Q, K_m)}_{\text{Music Alignment Drift}}
    \label{eq:forward}
\end{align}
with $\eta=1$ and $\nabla_{q_i}\mathcal{R}(Q, K_m)=-2\sum_j (S_m)_{i,j}\cdot q_j$ where $q_j$ is the $j$-th row vector of $Q$ and $\gamma_d$ is the hyperparameter. 
While the exact optimization yields the key matrix $K_t$, following standard Transformer architectures~\cite{ramsauerHopfieldNetworksAll2021}, we relax this by projecting the states into a separate value space $V_t$, enhancing the network's expressive capacity. 

While the theoretical music alignment drift $\nabla_Q \mathcal{R}$ provides a principled mechanism for music alignment, naively applying it within a diffusion framework introduces instabilities, such as magnitude explosion and high-frequency noise injection. To mitigate instability, we normalize music alignment drift term before applying it. Furthermore, we mask diagonal part of self-similarity matrix as it has strong bias signals from how they are computed (see at ~\cref{fig:ssm}). 

\subsubsection{Bayesian Update Module:}
\label{sec:MAP}
Because the text condition contains abstract, high-level information, it is underspecified early in the diffusion process~\cite{parkEnergyBasedCrossAttention2023}. During the generation process, as motion takes shape (\eg, a specific type of "happy jump dance"), the text embedding should be refined to focus on that specific mode of the distribution. Following previous work~\cite {parkEnergyBasedCrossAttention2023, EnergyMoGen}, we update the text conditions via MAP estimation. The gradient of the log-posterior is approximated as
\begin{align}
\nabla_{K_t} \log p(K_t|Q, K_m) = -(\nabla_{K_t} E(Q;K_t, K_m) + \nabla_{K_t} E(K_t))
\end{align}
where $E(K_t):=\text{lse}(\frac{1}{2}\text{diag}(K_tK_t^\intercal), 1)$.
We update the text embedding with $\alpha_t=0$:
\begin{align}
c_t^{i} = c_t^{i-1} + \Big(\delta_a\text{Softmax}(\frac{1}{\sqrt{d}} A)Q 
- \delta_r\mathcal{D}\big(\text{Softmax}(K_t')\big)K_t\Big)W_K^\intercal
\label{eq:update}
\end{align}
where $\mathcal{D}(\cdot)$ denotes diagonalization operators and $K_t'=\text{diag}(K_tK_t^\intercal)$ and $\delta_a, \delta_r$ are the hyperparameters to control update rate.

% This energy-driven update introduces three main advantages:  
% (1) \textit{Semantic consistency}—explicit alignment between motion and text;  
% (2) \textit{Iterative refinement}—MAP updates adapt embeddings across diffusion timesteps;  
% (3) \textit{Interpretability}—the energy landscape provides measurable alignment strength.  
% Overall, EBT transforms conventional attention into a learned inference process, guiding motion generation toward semantically coherent and musically synchronized outputs.

% \paragraph{Local and Global Attention Mask}
% Dance sequences may involve multiple textual cues within a single motion chunk, $x^{1:L}$, i.e., $c_{t}=(c_{text_1},c_{text_2}, c_{text_3})\in \mathbb{R}^{L\times D}$ each referring to distinct temporal intervals. To effectively model text-motion alignment, we introduce local attention masks. Local attention activates only time windows associated with the relevant text segment, enabling efficient, fine-grained control of motion semantics.
% This design parallels the multi-text attention strategy of GENMO~\cite{liGENMOGENeralistModel2025}, while maintaining computational efficiency.

\subsection{Loss Functions}
Our primary training objective follows the simplified DDPM formulation~\cite{ddpm}, in which the model estimates the clean motion sample~\cite{tevetHumanMotionDiffusion2022a} directly, rather than predicting noise~\cite{LDA} or velocity~\cite{mengAbsoluteCoordinatesMake2025}. The main loss is defined as
\begin{align}
    \mathcal{L}_{m}=\mathbb{E}_{x_0, t}[\|x_0 - \hat{x}_\theta(x_t, t, c_{t},c_{m})\|_2^2 ]
\end{align}
where $x_0\sim p(x_0|c_{t},c_{m}), t\sim [1, T]$. Following prior works~\cite{tevetHumanMotionDiffusion2022a, tseng2023edge, LDA}, we incorporate auxiliary objectives to improve physical plausibility and temporal smoothness, including joint position loss, foot-skating loss, and velocity loss. Further implementation details are provided in the supplementary material.
\section{Experiments}
\subsection{Experimental Settings}
\subsubsection{Datasets:}
We evaluate  \AlgName \ on AIST++~\cite{AIST} and \DataName \ (an extension of Motorica~\cite{LDA}) for dance motion generation. Because the two datasets use different human-body representations, we align Motorica to the  SMPL~\cite{SMPL} representation using LBFGS optimization.
As AIST++ doesn't have corresponding text information, we train the model only with music condition. 
% we extract prompts from DanceEditor~\cite{DanceEditor}, removing edit instructions but keeping descriptive text, yielding low-level descriptions for 44.3\% of the data. The dance genre serves as the high-level conditioning signal.

\begin{table}[t]
\centering
\caption{Comparison of selected single dance motion datasets. \textit{Text} refers to whether the dataset has corresponding text pairs. On the other hand, \textit{Dance Technique} is a domain-specific frame-level label, such as `Charleston cross step'.}
\resizebox{\columnwidth}{!}{
\begin{tabular}{l|c|c|c|c|c}
\Xhline{1pt}
\textbf{Dataset Name} & \textbf{Total hours} & \textbf{\# Genres} & \textbf{MoCap} & \textbf{Text} & \textbf{Dance Technique}   \\
\Xhline{1pt}
DanceRevolution~\cite{huangDanceRevolutionLongTerm2023} & 12 h & 3 & \x x & \x x & \x x \\
AIST++~\cite{aistpp} & 5.19 h & 10 & \x x & \x x & \x x \\
PopDanceSet~\cite{luoPOPDGPopular3D2024} & 3.56 h & 19 & \x x& \x x& \x x \\ 
DanceNet~\cite{M2D} & 0.96 h & 2 & \check c &\x x&\x x \\
ChoreoSpectrum3D~\cite{han2023enchantdance} & 70.32 h& 4 & \check c & \x x & \x x \\ 
Motorica Dance~\cite{LDA} & 6.22 h & 8 & \check c & \x x & \x x \\ 
FineDance~\cite{liFineDanceFinegrainedChoreography2023} & 14.6 h &  22 & \check c & \x x & \x x \\
DanceRemix~\cite{DanceEditor} & 117.39 h & 10 & \x x & \check c & \x x \\
\hline
\textbf{\DataName \ (Ours)} & \textbf{4.61 h} & \textbf{8} & \textbf{\check c}& \textbf{\check c}& \textbf{\check c} \\
\Xhline{1pt}
\end{tabular}%
}
    \label{tbl:dataset}
\end{table}

\subsubsection{\DataName:}
To improve text–motion alignment, we augment Motorica~\cite{LDA} with fine-grained dance annotations. Since dance motions differ significantly from general human movement (see ~\cref{fig:features}), a professional dancer labeled the Motorica frames by genre, dance technique, and detailed descriptions of body movements.
Due to missing musical references, 34 samples were excluded, leaving 97 fully annotated sequences, equivalent to 4.62 hours (see ~\cref{tbl:dataset}). To the best of our knowledge, this is the first dataset specifically annotated with frame-level domain-specific dance techniques and text-based motion descriptions. Please check the supplementary material for more details. 

\subsubsection{Evaluation Metrics:}
We follow traditional dance motion generation metrics, including both kinetic and geometric features, as proposed by ~\cite{AIST}. We report $\text{FID}_k$, $\text{FID}_g$, $\text{Dist}_k$, and $\text{Dist}_g$~\cite{FID, kinetic_feature, geometric_feature, siyaoBailando3DDance2022}, as well as the Beat Alignment Score (BAS)~\cite{AIST}. However, these metrics capture only the quality of the generated motion with respect to the music. Therefore, we propose a new experimental setup and metric to express dance editability.

\begin{table}[t]
    \centering
    \caption{Comparison with SoTAs on the AIST++ dataset and \DataName, where BAS refers to Beat-Alignment Score. $\rightarrow$ means closer to ground truth is better. `A' and `T' represent audio and text modality, respectively. \textbf{Bold} and \underline{underline} indicate the best and $2^{\text{nd}}$ results. TM2D* is trained with both audio and text.}
    \resizebox{\columnwidth}{!}{
    \begin{tabular}{l|l|c|c|c|c|c|c|c|c|c}
        \Xhline{1pt}
        & \multirow{2}{*}{\textbf{Method}}& \multirow{2}{*}{\textbf{Modality}} & \multicolumn{2}{c|}{\textbf{Motion Quality}} & \multicolumn{2}{c|}{\textbf{Motion Diversity}} & \multirow{2}{*}{\textbf{BAS $\uparrow$}} & \multicolumn{3}{c}{\textbf{EDS}} \\
        \cline{4-7}
        \cline{9-11} 
        & & &\textbf{$\text{FID}_k \downarrow$} & \textbf{$\text{FID}_g \downarrow$} & \textbf{$\text{Dist}_k \rightarrow$} & \textbf{$\text{Dist}_g \rightarrow$} & & \textbf{S$_{text}\uparrow$ }& \textbf{S$_{music}\uparrow$} & \textbf{EDS} $\uparrow$ \\
         \Xhline{1pt}
         \multirow{9}{*}{\rotatebox[origin=c]{90}{\textbf{AIST++}}}
         &Ground Truth & A & - & - & 10.03 & 7.38 & 0.2633  & - & - &- \\
         \cline{2-11} 
         &Li et al.~\cite{liLearningGenerateDiverse2020} & A &86.43  & 43.46  & 6.85  & 3.32  & 0.1607 & - & - &-  \\
         &DanceNet~\cite{M2D} & A &69.18  & 25.49  & 2.86  & 2.85  & 0.1430 & - & - &-  \\
        & DanceRevolution~\cite{huangDanceRevolutionLongTerm2023} & A &73.42  & 25.92  & 3.52  & 4.87  & 0.1950 & - & - &-  \\
         &FACT~\cite{aistpp} & A &35.35  & 22.11  & 5.94 & {6.18}  & 0.2209 & - & - &-  \\
         &Bailando~\cite{siyaoBailando3DDance2022} & A &\textbf{28.16}  & \textbf{9.62}  & \underline{7.83} & \underline{6.34}  & 0.2332 & - & - &-  \\
         &EDGE~\cite{tseng2023edge} & A &42.16  & 22.12  & 3.96 & 4.61  & \underline{0.2334}  & - & - &- \\
         &Lodge~\cite{lodge} & A & 37.09  & 18.79  & 5.58 & 4.85  & \textbf{0.2513} & - & - &- \\
         \cline{2-11} 
         &\AlgName \ (ours) & A & \underline{29.58} & \underline{11.54}  & \textbf{8.74} & \textbf{7.63} & {0.2312}  & - & - &-   \\
         \Xhline{1pt}
         \multirow{12}{*}{\rotatebox[origin=c]{90}{\textbf{\DataName\ (ours)}}}
         & Ground Truth & A + T & - & - & 10.54 & 7.33 & 0.2413  & - & - & - \\
         \cline{2-11}
         % & Lodge~\cite{lodge} & M &6595  & 17.96  &5.20  &4.95  & 0.282   \\
         & EDGE~\cite{tseng2023edge} & A  & 67.52 & 18.34 & 4.70 & 7.20 & 0.2052 &    0.8318 &0.3904 &0.5272 \\
         & POPDG~\cite{luoPOPDGPopular3D2024} & A & 27.02 & 8.56 & 6.73 & 5.80 &0.2345   & 0.8294 &\textbf{0.5365} & \underline{0.6501}  \\
         & DanceFusion~\cite{dancefusion} & A& 31.34 & 10.53  & 7.43 &7.33  &  0.2076 &0.8164    & 0.3383&0.4780 \\ 
        & Danceba ~\cite{danceba} & A & 41.24 &11.36  & 7.36  & 5.99 & 0.1947 &0.8363 &0.3983 & 0.5352\\
         % & TM2D*~\cite{TM2D}& A&123.11 &608.08&11.06& 16.44 &0.2611 &0.8748     & 0.5785     & 0.6933 \\
         \cline{2-11}
         % &\AlgName \ (ours) & A & &    &  & & & - & - &- \\
         % &\AlgName \ (ours) & A + T & 11.35 & 8.97 & 9.93 &6.68& 0.2491 & - & - &- \\
         % \Xcline{2-11}{1pt}
         & MDM~\cite{tevetHumanMotionDiffusion2022a} & T & 37.29 & 13.66  & 12.46 & 9.68 & -  & \underline{0.9684}   & \underline{0.4879}  & 0.6467 \\
         & MLD-5~\cite{chenExecutingYourCommands2023} & T & 87.51 &56.54 & 16.56 & 11.82 & - & \textbf{1.0}    & 0.3727      & 0.5423 \\ 
         & ReMoDiffuse~\cite{Remodiffuse}& T& 338.73 & 468.33  & 18.42 & 19.37 & -  & 0.8053  & 0.4860 & 0.6048\\
         & TM2D*~\cite{TM2D}& T& 313.08&308.58&18.38 &13.02 & -   &  0.8936  &0.4818  & 0.6252 \\
         \cline{2-11}
         & TM2D*~\cite{TM2D}& A + T& 126.46&570.84&\textbf{10.82} &15.23 & \textbf{0.2744}  & 0.8391   &0.4625     &0.5952 \\
         & UniMuMo~\cite{yangUniMuMoUnifiedText2025}& A + T& 28.10 & 169.14  &8.26 &11.06  & 0.2360 & 0.7514  & 0.3945 &0.5162 \\
         \cline{2-11}
         & \AlgName \ (ours) & A & 14.89 & 10.69 & 8.67 & \underline{7.01} &0.2249  & 0.8464  &0.4526 &0.5853 \\
         &\AlgName \ (ours) & T & 7.80 & \textbf{6.15} &9.99 & \textbf{7.08}  & - &\textbf{1.0}  &0.4533  & {0.6207} \\
         &\AlgName \ (ours) & A + T &  \textbf{7.32}  & \underline{6.87} & \underline{10.30} & 6.55 & \underline{0.2528} & \textbf{1.0}& 0.4870 & \textbf{0.6539}\\
         \Xhline{1pt}
    \end{tabular}
    }
    \label{tbl:main_table}
\end{table}

\subsubsection{Editable Dance Score (EDS):}
To properly assess whether the system can generate motion conditioned on text while faithfully following the given musical cues, we need to design a specialized experimental setup, which we call the \textit{Exchange Evaluation Protocol}. First, we select from the test dataset the samples longer than 3 seconds text-music pairs. Then we change the paired music to another based on BPM (Beat Per Minute) difference (we categorize three different tempo ranges- low, medium, and high). Lastly, we generate new text-music pairs, where the dance motion (from text) totally mismatches the music, \eg, fast hip hop dance with slow jazz music. 

To properly evaluate the exchange protocol, we propose a new metric,EDS, the harmonic mean of semantic preservation (S$_{text}$) and rhythmic adaptation (S$_{music}$). 
Specifically, we use the finetuned normalized TMR~\cite{petrovich23tmr} CLIP score as S$_{text}$ and normalized BAS as S$_{music}$, then we define EDS as $\text{EDS}:= \frac{2\cdot \text{S}_{text}\cdot\text{S}_{music}}{\text{S}_{text}+\text{S}_{music}}$. EDS can capture whether the generated motions are semantically and rhythmically well-aligned. For more details, please check supplementary material.

\subsubsection{Implementation Details:}
All datasets are segmented into 5-second clips with a 1-second hop size and sampled at 30~FPS for both motion and audio, yielding sequences of length $L=150$.
For long-term dance generation, we stitch 2.5-second segments in an overlapping manner using the dynamics CFG and latent blending, similar to EDGE~\cite{tseng2023edge} (please see the supplementary material for more details).
The model employs 9 BEAM layers with an embedding dimension of $D=512$. 

\subsubsection{Baselines:}
Since we propose a 'text + music to motion generation pipeline' with a new dataset, we retrain state-of-the-art dance models~\cite{tseng2023edge, luoPOPDGPopular3D2024,dancefusion}, text to motion models~\cite{tevetHumanMotionDiffusion2022a,chenExecutingYourCommands2023,Remodiffuse}, and text+music-conditioned models~\cite{TM2D,yangUniMuMoUnifiedText2025}, following their original training recipes. 
% As the most dance generation architectures are trained and evaluated with music chunk, while text-motion architectures are based on text-motion pairs, we report diff

\begin{figure*}[t]
	\centering
	\includegraphics[width=\linewidth]{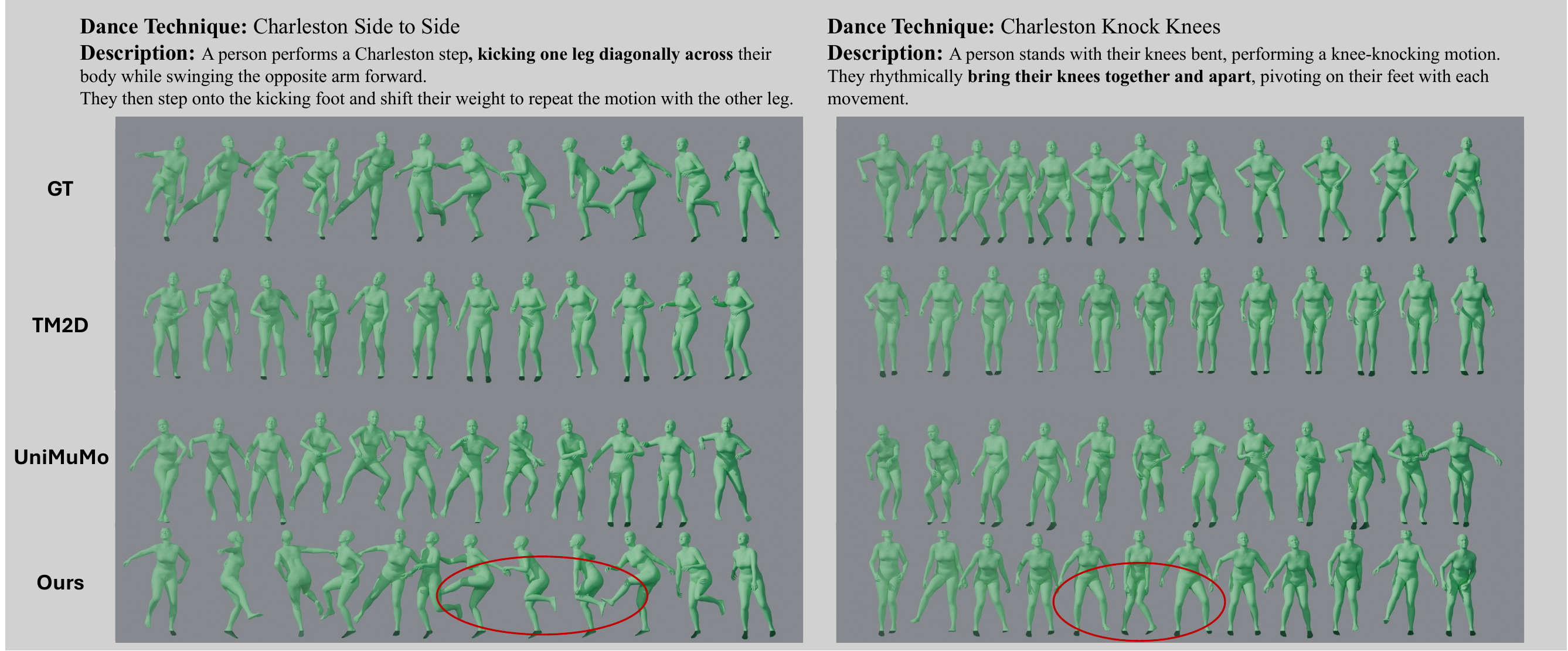}
    % \vspace{-3mm}
	\caption{
	    Qualitative dance motion generation results with text and music conditioned, compared with other SoTA models. 
	}
	\label{fig:quality}
    % \vspace{-3mm}
\end{figure*}

\begin{figure*}[t]
	\centering
	\includegraphics[width=\linewidth]{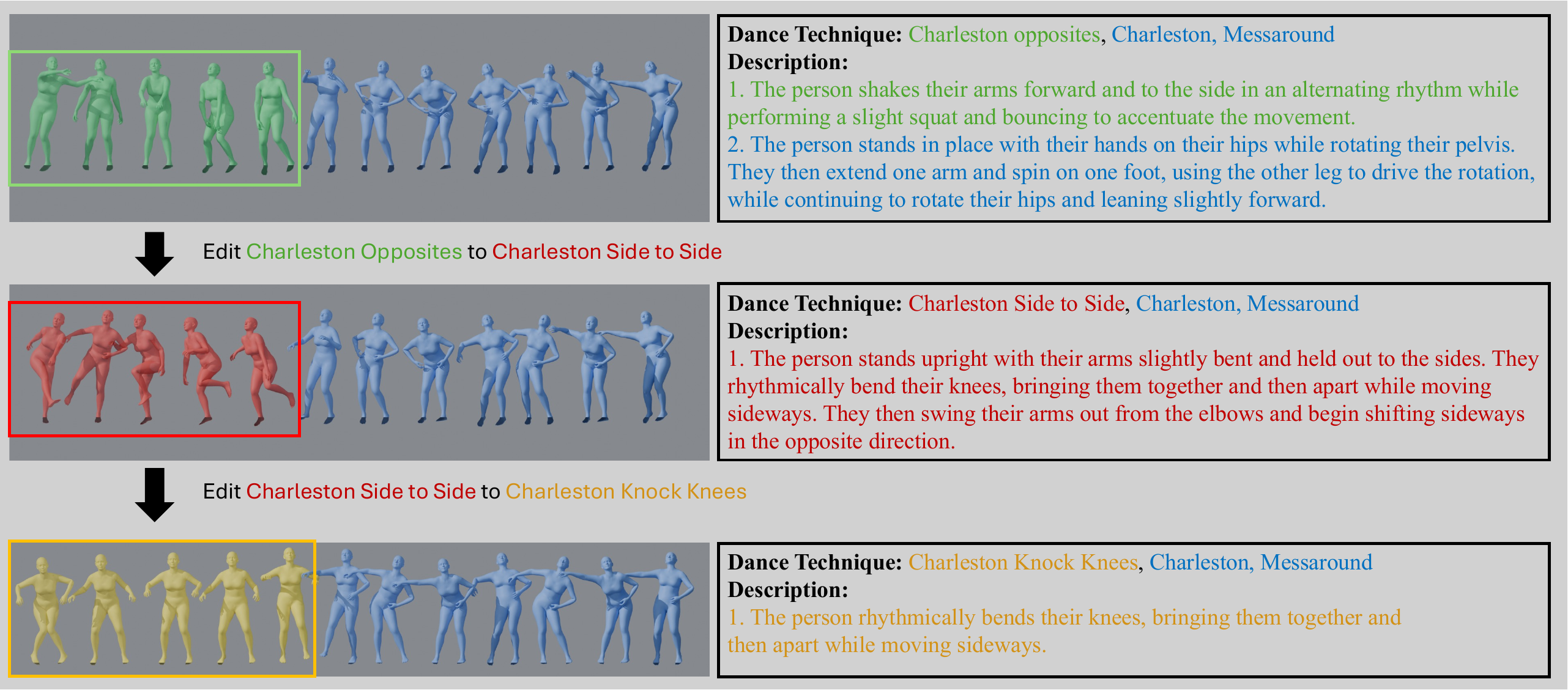}
	\caption{
	     Visualization of dance motion editing example. The original motion contains two dance techniques: Charleston Opposites (green) and Charleston Messaround (blue). We first edit Charleston Opposites to Charleston Side to Side (red) while preserving the other. Similarly, we can edit one more time to Charleston Knock Knees (yellow).
	}
	\label{fig:edit}
\end{figure*}

\subsection{Evaluation of Dance Motion Generation}
We evaluate \AlgName \ on two datasets: 1. AIST++~\cite{aistpp}, and 2. \DataName. Since, AIST++ doesn't have text annotations, we train our model using only music conditioning and evaluate it with standard metrics such as FID, Dist, and BAS. 
In contrast, \DataName \ contains high-quality text annotations. Therefore, we train both music-to-motion and text-to-motion models, including models conditioned on both music and text. Because, most text-motion models are trained and evaluated using text-motion pairs (with motion length varying depending on the text prompt), we evaluate the entire \DataName \ based on text-motion pairs, where motion durations range from 0.5 seconds to 10 seconds.
Quantitative results are reported in ~\cref{tbl:main_table}. 
On \DataName, our method achieves state-of-the-art performance across FID, Dist, and BAS. On AIST++, \AlgName \ attains the best motion diversity (Dist) while remaining competitive with specialized music-to-dance models on FID and BAS. 
% On both the AIST++ and \DataName\ benchmarks, our method achieves state-of-the-art performance on FID, Dist, and BAS compared to existing pipelines.
~\cref{fig:quality} presents qualitative comparisons with SoTA methods~\cite{TM2D, yangUniMuMoUnifiedText2025} under text conditioning. The results show that our model can generate sophisticated dance motions that closely follow the provided text prompts. 

\subsection{Evaluation of Editable Dance Generation}
% \subsubsection{Quantitative Results:}
We evaluate the editability of dance motion generation using the proposed \textit{Exchange Evaluation Protocol} with EDS metric in ~\cref{tbl:main_table}, which measures the joint alignment performance w.r.t. both text and music.
Most single-modality methods exhibit a trade-off between $\text{S}_{music}$ and $\text{S}_{text}$. For example, POPDG achieves (0.8294 vs 0.5365) and MLD (1.0 vs 0.3727), indicating strong performance in one modality but weaker alignment in the other.
Furthermore, existing audio-text multimodal pipelines often fail to capture both modalities effectively due to the way the two modalities are combined. 
For instance, TM2D~\cite{TM2D} primarily uses audio features and incorporates text features through window-based late fusion. Because of this design, the features are conflicted in some samples, showing TM2D text-only shows better result.
In contrast, \AlgName \ disentangles text and music modalities through the proposed energy function and Hierarchical CFG. As result, the model achieves strong text controllability while maintaining accurate musical alignment, enabling effective editing of dance motions based on text prompts. \cref{fig:edit} shows qualitative results of editing quality. The initial dance text conditions are \textit{Opposites} and \textit{Messaround}. We edit only Opposites to \textit{Side to Side} or \textit{Knock Knees}. This demonstrates fine-grained, text-conditioned edits with high motion fidelity.

\subsection{Ablation Study}
\label{sec:abl}
\subsubsection{Proposed Modules:}
We conduct ablation studies to evaluate the effectiveness of the proposed modules by varying $\gamma_m$ and $\gamma_d$ in the energy function in ~\cref{eq:forward}, as well as the normalization applied to the music alignment drift term ($\nabla_Q \mathcal{R}$). Compared with the Abl-4 model in ~\cref{tbl:abl}, removing the music alignment energy function mainly reduces the beat alignment score (BAS). Although other quality metrics remain largely unchanged, the BAS decreases, indicating that the model relies more heavily on textual information.
Similarly, removing AdaLN leads to a significant drop in BAS. This suggests that the model attempts to capture global structure from the music features, placing excessive reliance on them and limiting its ability to capture fine-grained local beat information. 
Finally, removing the normalization at $\nabla_Q \mathcal{R}$ degrades both the overall generation quality and the BAS.

\subsubsection{Music Context Update:}
\label{sec:Music_update}
Similar to text context Bayesian update at ~\cref{eq: update}, one could theoretically update the music condition via MAP estimation:$c_m^i=c_m^{i-1} - \nabla_{K_m}\big(E(Q;K_t, K_m) +E(K_m)\big)$. 
However, we empirically find this symmetric update degrades performance. While the text condition provides abstract, global semantics that benefit from iterative refinement as the motion materializes, the music condition provides strict, fine-grained temporal anchors. If the music context is updated via MAP estimation, the model tends to minimize the energy landscape by shifting the music condition to align with the currently generated motion, rather than forcing the motion to adapt to the true musical beats. Consequently, the context slowly deviates from the ground-truth acoustic properties, deteriorating both overall motion generation quality and Beat Alignment Scores (BAS), as demonstrated in ~\cref{tbl:abl}. Therefore, \AlgName \ employs an asymmetric update strategy: iteratively refining the abstract text semantics while keeping the rhythmic music condition strictly frozen.
\begin{table}[t]
    \centering
    \caption{Ablation study on different $\gamma$ and normalization at ~\cref{eq:forward}, AdaLN and Bayesian music context $c_m$ update. Note that $\gamma_m=\gamma_d=0$ means the model only uses text information.}
    % \resizebox{\columnwidth}{!}{
    \begin{tabular}{l|c|c|c|c|c}
        \Xhline{1pt}
        \textbf{Name} & \textbf{$\text{FID}_k \downarrow$} & \textbf{$\text{FID}_g \downarrow$} & \textbf{$\text{Dist}_k \rightarrow$} & \textbf{$\text{Dist}_g \rightarrow$} & \textbf{BAS} $\uparrow$ \\ 
         \Xhline{1pt}
         \AlgName ($\gamma_m=1.0, \gamma_d =1.0$) & 7.32 &6.87&10.30&6.55 & 0.2528   \\
         \hline
         Abl-1 ($\gamma_m=1.0, \gamma_d =0.5$)  & 9.33 & 5.31 & 10.79 &6.42 &0.2405 \\
         Abl-2 ($\gamma_m=1.0, \gamma_d =0.0$)  & 9.10 &6.19 & 10.21 &6.38 & 0.2369 \\
         Abl-3 ($\gamma_m=0.0, \gamma_d =0.5$)  & 8.32 & 6.58& 10.60& 6.43&0.2468 \\
         Abl-4 ($\gamma_m=0.0, \gamma_d =0.0$)  &10.38 & 5.75 &9.04 &6.38 &0.2301   \\
         \hline
         w/o Norm ($\gamma_m=1.0, \gamma_d =0.5$) & 10.37 & 10.17& 10.34& 7.30 & 0.2251   \\
         \hline
         w/o AdaLN ($\gamma_m=1.0, \gamma_d =1.0$) & 7.79& 4.28 & 10.54& 7.33 &0.2285  \\
         \hline
         $c_m$ update  &7.16 & 7.32 &10.51 &6.58 &0.2372  \\
         \Xhline{1pt}
    \end{tabular}
    % }
    \label{tbl:abl}
\end{table}

\section{Conclusion and Limitation}
In this work, we present \AlgName, a Structural-Temporal Rhythmic Energy-based Attention for dance Motion generation pipeline that jointly leverages music and text conditions via a novel energy function. By disentangling these modalities, text conditions govern the global motion structure and semantics of the motion, while music conditions enrich local temporal details, enabling controlled and expressive dance synthesis. 
In addition, we introduce \DataName, a  dance-specialized text-motion dataset that enables the generation of fine-grained dance techniques through language descriptions. 
% A current limitation of proposed EDS is based on BAS to compute music alignment. However, BAS acts as a recall metric (measuring the distance from music beats to the nearest kinematic beat), models that lack rhythmic prior 
A current limitation of our system is its focus on single-person dance generation. Many choreographic scenarios involve group performance, where dancers exhibit complex spatial formations, coordinated timing, and inter-person interactions. Expanding \AlgName\ to multi-dancer settings thus represents an important direction for future work.

\section{Acknowledgements}
We gratefully acknowledge support from the National Science Foundation under Grant BCS-2318255 and from the University of Maryland through the AIM Research Seed Award Program.

% \section*{Acknowledgements}
% Please insert your acknowledgments here.

\bibliographystyle{splncs04}
\bibliography{main}
\clearpage
% \Large
\textbf{Supplementary Material: Text Dictates, Music Decorates: Energy-based Attention for Editable Dance Motion Generation}\\
\vspace{1.0em}

In the supplementary material, we provide additional details including extended related works (Sec.~\ref{sec:extended_related_works}), details of \DataName (Sec.~\ref{sec:data}), details of method (Sec.~\ref{sec:method}), experiments results (Sec.~\ref{sec:exp}), and Dance Design Studio (Sec.~\ref{sec:studio}).  For more qualitative results, please check the videos. 

\section{Further Related Works}
\label{sec:extended_related_works}
\subsection{Datasets for Human Motion and Dance Generation}
The quality of dance datasets, in terms of motion accuracy and synchronization with music, closely influences the quality of the generated dance motion.
High quality datasets reduce self-penetration and contribute to more visually plausible results. Several datasets have supported progress in synthetic dance generation. 
AIST++ \cite{aistpp} provides approximately 5.2 hours of dance across 10 genres and is widely used in dance motion generation research. However, because AIST++ was recorded without a marker-based Motion Capture (MoCap) system, its motion fidelity is lower than that of more recent datasets.  ChoreoSpectrum3D \cite{han2023enchantdance} with EnchantDance, offers 70.32 hours of motion across four coarse genres (Pop, Ballet, Latin, House) and it is recorded with marker-based MoCap, providing higher-quality sequences. Similarly, the Motorica dance dataset~\cite{LDA} is captured with marker-based MoCap and covers a diverse set of genres. Despite these advances, most existing dance datasets only contain music-motion pairs, which limits the level of controllable generation available to choreographers. 
On the other hand, general human motion text datasets such as HumanML3D~\cite{HumanML3D}, KIT-ML~\cite{KIT} and Motion-X~\cite{linMotionXLargescale3D2023a} contain dense motion-text annotations. 
However, as discussed in the Introduction (Sec.~\ref{sec:intro}), general human motion and dance motion differ substantially in dynamics (see also Fig.~\ref{fig:features}). 
Consequently, even when models are trained jointly  on text-motion and dance motion datasets, they struggle to generate controllable dance motions from text prompts, as shown in Fig.~\ref{fig:quality}. 

Recently, \cite{DanceEditor} proposed a dance-and-text dataset, extended from AIST++~\cite{aistpp}. They retrieve similar dance motion pairs and generate text prompts using LLMs. Although DanceRemix provides a text-conditioned dance motion generation baseline, it does not include domain-specific textual information such as dance techniques.

\section{\DataName}
\label{sec:data}
To the best of our knowledge, \DataName \ is the first single dance dataset annotated by a professional dancer with detailed, expert-level labels. The most comparable datasets are DanceRemix~\cite{DanceEditor} and MDD~\cite{Gupta_2025_ICCV}. However, DanceRemix concentrates primarily on editing one dance motion into another, and MDD focuses on duet dance motions.

\begin{figure}[t]
	\centering
	\includegraphics[width=\linewidth]{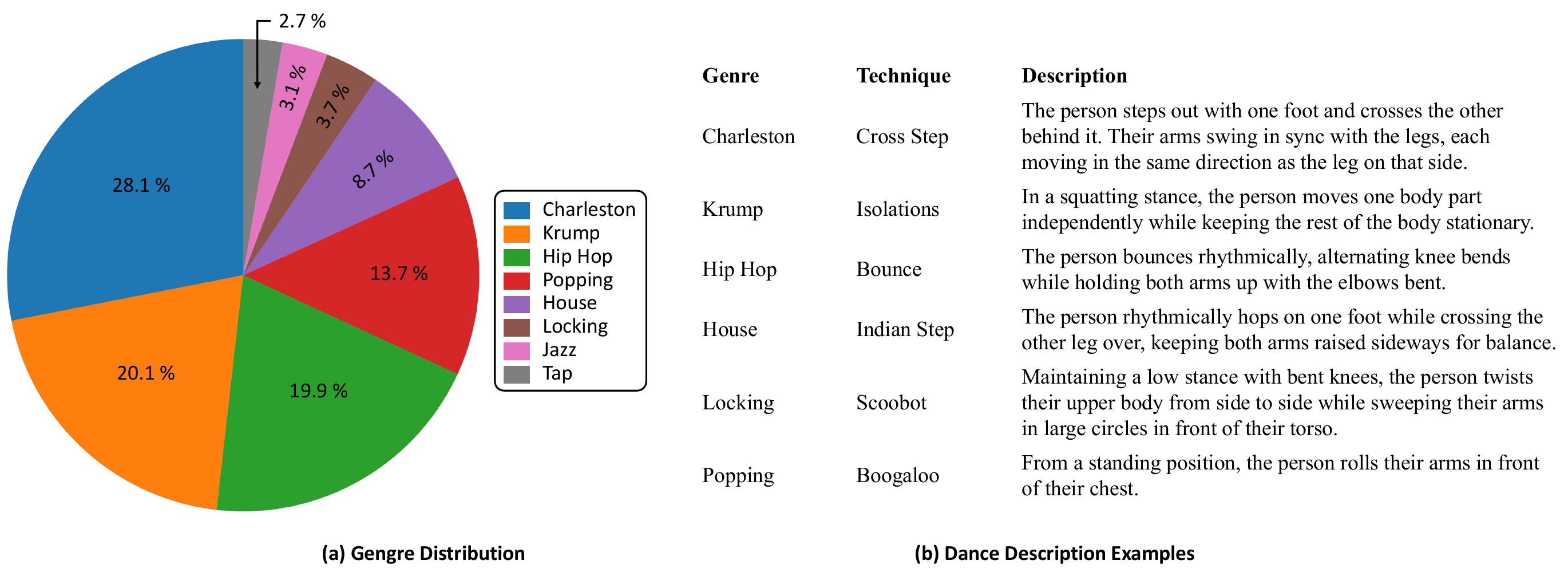}
	\caption{
	    Genre distribution and \DataName \ dance description examples.
	}
	\label{fig:dataset}
\end{figure}

On the other hand, \DataName \ is annotated with a total of 183 dance techniques and their corresponding descriptions (Fig.~\ref{fig:dataset}). While the Motorica dance dataset labels data only at the genre level based on music, our dataset provides fine-grained annotations of both genres and techniques, \ie, distinguishing different movements even within sequences sharing the same music. For example, the sample \textit{kthstreet-gPO-sFM-cAll-d02-mPO-ch01-bombom-001} is labeled as \textit{popping} in the Motorica dance dataset, but it actually contains four distinct genre-specific dance techniques in our annotations.

\begin{figure}[t]
	\centering
	\includegraphics[width=\linewidth]{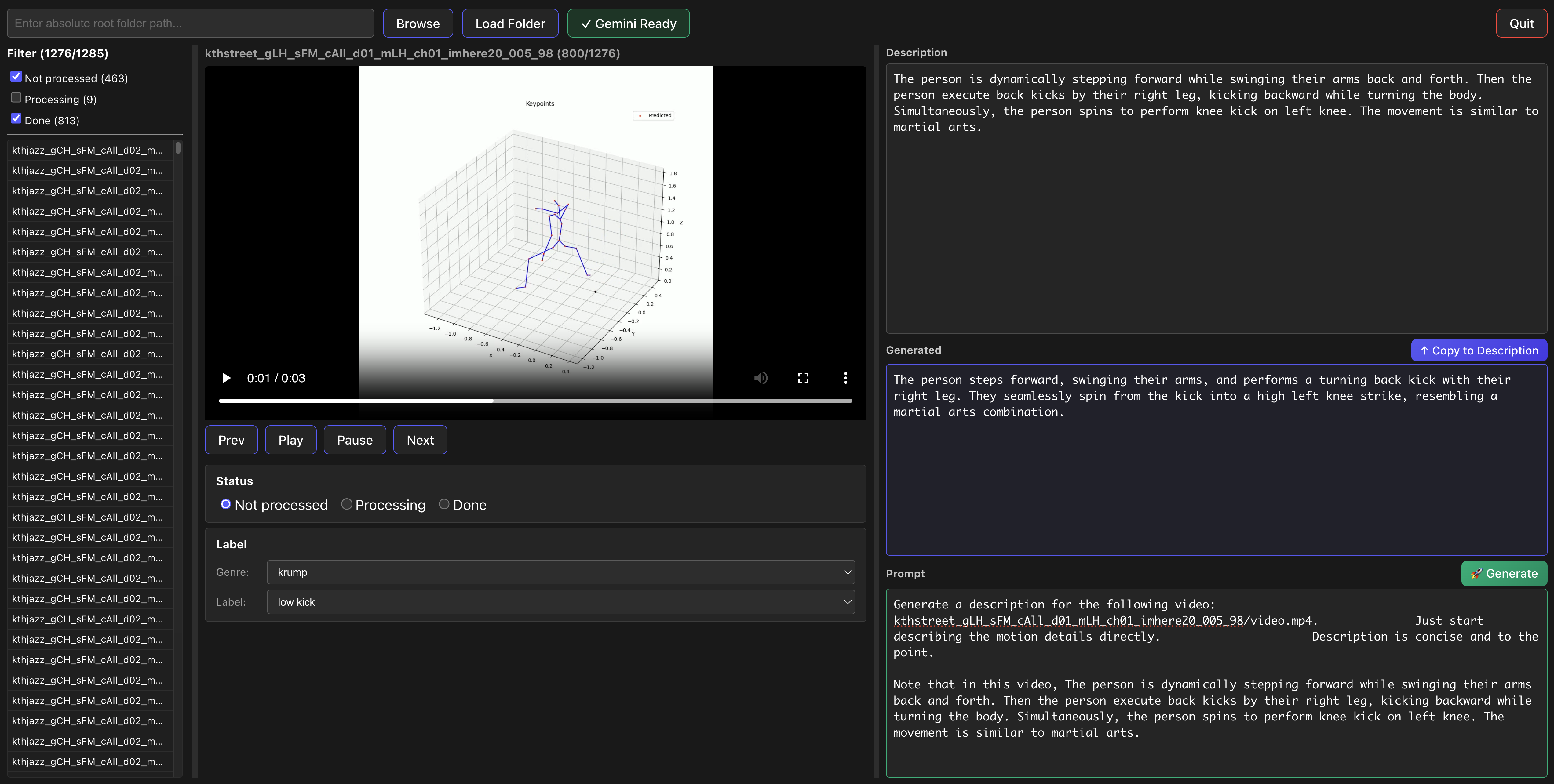}
    % \vspace{-3mm}
	\caption{
	    Custom video description correction tool. On the right window, we can edit the originally generated description, or we can regenerate by editing the prompt.
	}
	\label{fig:video_cap}
\end{figure}

After annotating the dataset, we further enhance the quality of \DataName \ by adding detailed text descriptions. First, we generate reference descriptions based on genres and dance techniques labels. Next, we use Gemini with sliced video clips, genres, label, and reference text to generate dance-specific human motion explanations. Then, we filter out samples that are not properly aligned with the generated text based on CLIP scores of a pretrained TMR~\cite{petrovich23tmr} ($\leq 0.7$). Finally, we manually correct these descriptions using the custom video annotation tool shown in Fig.~\ref{fig:video_cap}, focusing specifically on how the human body moves within the video clip. Consequently, \DataName \ contains both specific dance labels and detailed text explanations of human motion, making it compatible with other text-motion datasets. 

\section{Method}
\label{sec:method}
\subsection{Dual Energy-Based Cross Attention (D-EBCA)}
In the main manuscript, we propose a new energy function at D-EBCA module(~\cref{sec:DEBCA}), which is  
\begin{align}
E(Q;K_t,K_m) &= - \frac{1}{\beta}\sum_{l=1}^L \text{lse}(A_l, \beta) + \mathcal{R}(Q, K_m) \\ \nonumber
& +\frac{\alpha_t}{2}\|K_t\|^2 +\frac{\alpha_m}{2}\|K_m\|^2 +\frac{1}{2}\|Q\|^2
\label{eq:main_energy_app}\\
    \mathcal{R}(Q, K_m)&=-\sum_{i,j}(S_m)_{i,j}\cdot(q_i^\intercal q_j)
\end{align}
where 
\begin{align}
Q &= xW_{Q}, \quad K_t = c_tW_{K^t}, \quad K_m=c_mW_{K^m}, \quad V = c_tW_{V} \\ 
A&=QK_t^\intercal + \gamma_m Q K_m^\intercal \\ 
S_m&=\hat{K}_m \hat{K}_m^\intercal
\end{align}
with $\hat{K}_m$ is row-normalized $K_m$, $A_{l}$ is $l$-th row vector of $A$, $x\in\mathbb{R}^{L\times D_x}$ is motion latent, $c_t\in \mathbb{R}^{L_t \times D_t}$ and $c_m\in\mathbb{R}^{L_m\times D_m}$ are text and music conditions, respectively. In practice, we use $L=L_t=L_m=150$ and $D_x=D_t=D_m=512$. $S_m$ is the self-similarity matrix~\cite{SSM} of $K_m$. 
$\mathcal{R}(Q, K_m)$ is the music alignment energy function, measuring how well musical information is aligned with the given motion query.

Minimizing ~\cref{eq:main_energy_app} using the CCCP yields the forward path:
\begin{align}
    Q_{new}&=Q - \eta\nabla_Q E(Q, K_t, K_m)\\
    &=Q - \eta\left(Q-\nabla_Q\frac{1}{\beta}\sum_{l=1}^L \text{lse}(A_l, \beta) +\nabla_Q \mathcal{R}(Q, K_m)\right) \\ 
    &= (1-\eta)Q+\eta\nabla_Q\frac{1}{\beta}\sum_{l=1}^L \text{lse}(A_l, \beta) -\eta\nabla_Q \mathcal{R}(Q, K_m)
    \label{eq:last_forward_term}
\end{align}

Then computing gradient with respect to a single row vector of $Q$ is following:
\begin{align}
\nabla_{q_i}\frac{1}{\beta}\sum_{l=1}^L \text{lse}(A_l, \beta ) &= \frac{1}{\beta}\nabla_{q_i} \log \sum_{j=1}^L \exp\left(\beta (q_i(k_t)_j^\intercal + \gamma_m q_i (k_m)_j^\intercal\right) \\
&=\frac{1}{\beta}\sum_{j=1}^L \frac{\exp\left(\beta (q_i (k_t)_j^\intercal+\gamma_mq_i(k_{m})^\intercal_j)\right)}{\sum_{k=1}^L \exp\left(\beta q_i (k_t)_k^\intercal + \gamma_m q_i(k_m)_k^\intercal\right)}\nabla_{q_i}(\beta A_{i,j}) \\ 
&=\text{Softmax}(\beta A_i)(K_t + \gamma_m K_m)
\end{align}

Therefore 
\begin{align}
\nabla_Q \sum_{l=1}^L \text{lse}(A_l,\beta )&= \text{Softmax}\left(\beta A\right)(K_t + \gamma_m K_m) \\
&\approx\text{Softmax}\left(\frac{1}{\sqrt{d}}A\right)V_t.
\end{align}

The last approximation is achieved by setting $\beta=1/\sqrt{d}$ and replace $K_t+\gamma_m K_m$ to $V_t$, following the previous works~\cite{ramsauerHopfieldNetworksAll2021}. 

Then finally, ~\cref{eq:last_forward_term} becomes ~\cref{eq:forward}:
\begin{align}
Q_{new}=\underbrace{\text{Softmax}\left(\frac{1}{\sqrt{d}}A\right)V_t}_{\text{Dual Attractor}} - \gamma_d \underbrace{\nabla_Q \mathcal{R}(Q, K_m)}_{\text{Music Alignment Drift}}
\end{align}
where $\nabla_Q\mathcal{R} (Q, K_m)=-2 S_m Q$ and $\eta=1$. 
\begin{table}[t]
    \centering
    % \vspace{-3mm}
    \caption{Ablation study of different $\delta_a$ and $\delta_r$ at Eq.~\ref{eq:update}. All experiments trained on \DataName. \textbf{Bold} and \underline{underline} indicate the best and $2^{\text{nd}}$ results. TM2D* is trained with both audio and text.}
    % \resizebox{\columnwidth}{!}{
    \begin{tabular}{l|c|c|c|c|c}
        \Xhline{1pt}
        \multirow{2}{*}{\textbf{Method}}&\multicolumn{2}{c|}{\textbf{Motion Quality}} & \multicolumn{2}{c|}{\textbf{Motion Diversity}}& \multirow{2}{*}{\textbf{BAS $\uparrow$}}\\
        \cline{2-5} 
         & \textbf{$\text{FID}_k \downarrow$} & \textbf{$\text{FID}_g \downarrow$} & \textbf{$\text{Dist}_k \rightarrow$} & \textbf{$\text{Dist}_g \rightarrow$} &  \\
         \Xhline{1pt}
         Ground truth & - & - & 10.54 & 7.33 & 0.2413 \\
         \hline
         $\delta_a=0.1, \delta_r =0.1$ & 50.40& 356.80& 12.45& 17.78 & 0.2447\\
         $\delta_a=0.05, \delta_r =0.05$ &11.85& \underline{7.45}& 9.21 & 6.18 & 0.2396 \\
         $\delta_a=0.01, \delta_r =0.01$ &{7.95} &8.95 & 10.03& {6.62} & 0.2446 \\
         $\delta_a=0.005, \delta_r =0.005$ & 8.57 &7.53 &\textbf{10.49}  & \underline{6.69} &\underline{0.2462} \\
         $\delta_a=0.002, \delta_r =0.002$ &\underline{7.32} & \textbf{6.87}&{10.30} &6.55 &\textbf{0.2528} \\
         $\delta_a=0.0, \delta_r =0.0$ & \textbf{6.75}& 10.73 &\underline{10.46}&\textbf{6.98} &0.2444 \\
         \Xhline{1pt}
    \end{tabular}
    % }
    % \vspace{-3mm}
    \label{tbl:delta}
\end{table}

\subsection{Detailed Loss Function}
The main loss is defined as
\begin{align}
    \mathcal{L}_{m}=\mathbb{E}_{x_0, t}[\|x_0 - \hat{x}_\theta(x_t, t, c_{t},c_{m})\|_2^2 ]
\end{align}
where $x_0\sim p(x_0|c_{t},c_{m}), t\sim [1, T]$. Following prior works~\cite{tevetHumanMotionDiffusion2022a, tseng2023edge, LDA}, we incorporate auxiliary objectives to improve physical plausibility and temporal smoothness, including joint position loss ($\mathcal{L}_j$), foot-skating loss ($\mathcal{L}_f$), and velocity loss ($\mathcal{L}_v$) as follows:

\begin{align}
    \mathcal{L}_j &= \frac{1}{N}\sum_{i=1}^N \| \text{FK}(x_0^i) - \text{FK}(\hat{x}_\theta^i) \|_2^2 \\ 
    \mathcal{L}_{f} &= \frac{1}{N-1} \|(\text{FK}(\hat{x}_{\theta}^{i+1}) - \text{FK}(\hat{x}_\theta^{i}))\cdot f^i  \|_2^2 \\
    \mathcal{L}_{v}&= \frac{1}{N-1} \|(\text{FK}(x_0^{i+1})-\text{FK}(x_0^i))-(\text{FK}(\hat{x}_\theta^{i+1})-\text{FK}(\hat{x}_\theta^i)) \|_2 ^2 
\end{align}

where $\text{FK}(\cdot)$ is forward kinematics and $f^i\in\{0, 1\}^J$ is foot joint mask.

\subsection{Long Range Generation}
To generate motion sequences beyond the fixed 5-second training window, we adopt a sliding-window overlapping approach with 2.5 second stride between consecutive chunks~\cite{tseng2023edge}. At each chunk boundary, we condition the diffusion reverse process on the previously generated motion using three mechanisms. First, we construct a temporal blend mask that hard-constrains the overlapping prefix frames and linearly ramps to zero over a short blend zone, ensuring that the known past is preserved while the new segment is freely generated. 
Second, we apply temporally-varying classifier-free guidance, ramping the text guidance scale from zero in the prefix region to its full value in the generation region. This prevents the text condition from re-initializing the semantic content at each chunk, which would otherwise produce discontinuous style transitions.
Third, at each denoising step, we inject the clean prefix into the current noisy latent at the appropriate noise level using the DDPM forward process with a fixed noise realization, blending only the rotation channels while leaving root translation unconstrained. 
The final long-range sequence is assembled by aligning consecutive chunks via forward kinematics based rotation matching and SLERP interpolation at the boundaries. 

\begin{figure}[t]
    \centering
  \includegraphics[width=\linewidth]{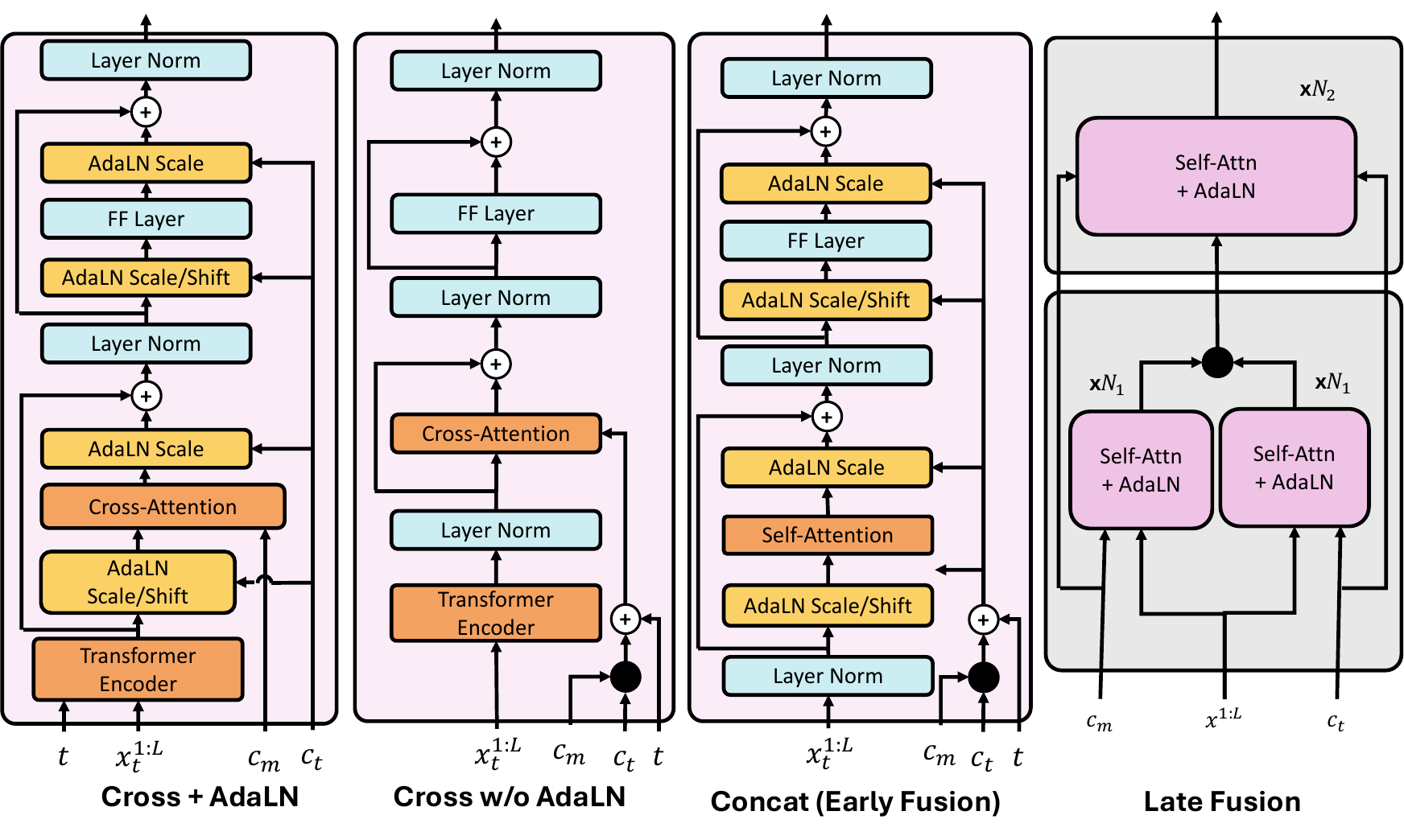}
   \caption{Detailed Ablation architecture diagrams. Late Fusion is represented multiple layers, while others show one details of each layer. Black dots represents concatenation. For Late Fusion, we use $N_1 =0.7 N$ and $N_2=0.3N$, where $N$ is total number of layers.}
   \label{fig:abl_arch}
\end{figure}

\section{Further Experiments Results}
\label{sec:exp}
\subsection{Ablation Study}
We conduct an ablation study on the effects of text context Bayesian update described at ~\cref{sec:MAP}. As shown in ~\cref{tbl:delta}, 
updating the text context with MAP estimation is effective~\cite{EnergyMoGen, parkEnergyBasedCrossAttention2023}. However, increasing the update hyperparamters not only lowers the beat alignment scores (BAS) but also deteriorates motion generation quality. We hypothesize that applying MAP estimation at early stage of diffusion reverse process could deviate text condition to much from the ideal manifold. 

 Furthermore, we design and compare multiple standard variations of multimodal architectures (\cref{fig:abl_arch}), and results are reported in ~\cref{tbl:abl_5}. Standard Cross-Attention also performs poorly on generation and Exchange experiments (EDS), consistent with our modality collapse argument. 
Late Fusion achieves comparable BAS and $\text{FID}_g$ but degrades in $\text{FID}_k$ and $\text{EDS}$, showing that  \AlgName \ more effectively combines text and audio modalities. 
Note that $\text{S}_{\text{text}}=1$ represents the distribution ceiling, not the metric defect; differentiation is preserved across architectures.

\begin{table}[b]
    \centering
    \caption{Extended ablation study on different architectures trained and evaluated on \DataName.}
    \resizebox{\columnwidth}{!}{
    \begin{tabular}{l|c|c|c|c|c|c|c|c|c}
        \Xhline{1pt}
        \textbf{Name} & \textbf{\# Params} & \textbf{$\text{FID}_k \downarrow$} & \textbf{$\text{FID}_g \downarrow$} & \textbf{$\text{Dist}_k \rightarrow$} & \textbf{$\text{Dist}_g \rightarrow$} & \textbf{BAS} $\uparrow$ &\textbf{S$_{text}\uparrow$ }& \textbf{S$_{music}\uparrow$} & \textbf{EDS} $\uparrow$ \\
         \Xhline{1pt}
         \AlgName & 91.9 M & \textbf{7.32} &\textbf{6.87}&\textbf{10.30}&\textbf{6.55} & \textbf{0.2528}   & \textbf{1.0} & \textbf{0.4870} & \textbf{0.6539}  \\
         \hline
         Cross-Atten + AdaLN &82.1 M & 236.88 & 487.96 & 13.79 & 15.37 & 0.2327  &0.9488  & 0.4018   & 0.5641 \\
         Cross-Atten + w/o AdaLN & 72.2 M &47.93 & 11.23 &4.99 &6.23 & 0.1980 &0.8553 &0.3235  &0.4674  \\
         Concat (Early Fusion) & 78.5 M & 26.72 & 178.28 & 8.22 & 8.44 & 0.2372 & \underline{0.9507} & 0.4220 & 0.5829 \\
         Late Fusion & 88.7 M & 20.22 & 10.11 & 6.82 & 5.45 & \underline{0.2497} & 0.9491 & 0.4272 & 0.5887\\
         \hline
         \AlgName \ (HumanML3D + \DataName) & 91.9 M & \underline{7.63} & \underline{8.89} & \underline{9.87} & \underline{6.95} & 0.2440 & \textbf{1.0} & \underline{0.4784} & \underline{0.6456} \\
          \Xhline{1pt}
    \end{tabular}
    }
    \label{tbl:abl_5}
\end{table}

\begin{figure}[t]
    \centering
    \includegraphics[width=\textwidth]{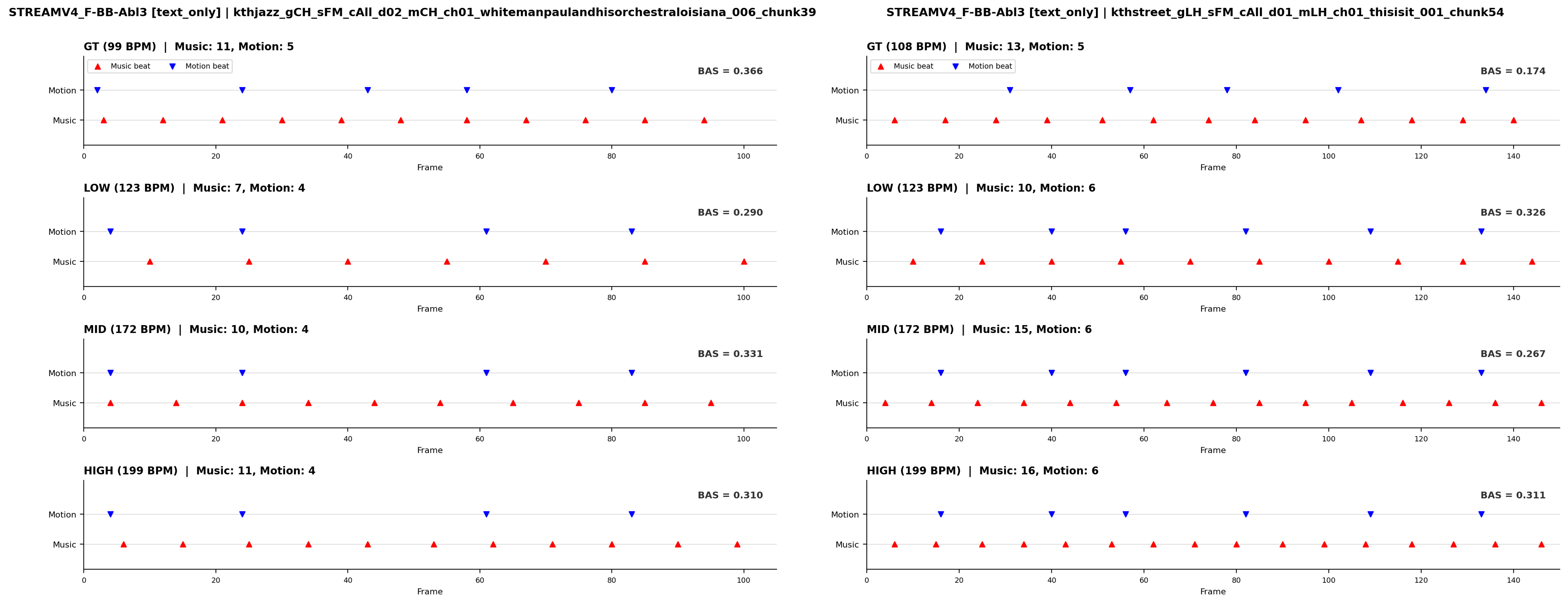}
    \caption{Visualization of music beats (red) and motion beats (blue) generated by the text-only STREAM model. Since the model only uses text information, it lacks the ability to adapt motion to different music. However, because dance motion inherently contains natural motion beats, the BAS can still be high even when paired with different music.}
    \label{fig:eds}
\end{figure}
\subsection{Editable Dance Score}
To properly evaluate the exchange protocol, we propose a new metric, EDS, defined as the harmonic mean of semantic preservation (S$_{text}$) and rhythmic adaptation (S$_{music}$). 
Specifically, we use the finetuned normalized TMR~\cite{petrovich23tmr} CLIP score as S$_{text}=\frac{C_{text}^{recon}}{\|C_{text}^{gt}\|}$, where $C_{text}^{recon}$ and $C_{text}^{gt}$ denote the CLIP scores of the reconstructed and ground truth motions, respectively. 
Because the model is trained on a dance dataset, the generated motion naturally contains motion beats even without music conditioning, as illustrated in ~\cref{fig:eds}. 
Therefore, relying solely on the BAS score cannot reliably determine whether the generated motion truly adapts to the given musical beats. 
Therefore, we apply a correction weight ($W_{beat}:=\frac{\text{\# motion beats}}{\text{\# music beat}}$) to the normalized music score, yielding S$_{music}=W_{beat}\frac{\text{BAS}^{recon}}{\text{BAS}^{gt}}$.
Based on the normalized text score and music score, the final definition of EDS is
\begin{align}
    \text{EDS}:= \frac{2\cdot \text{S}_{text}\cdot\text{S}_{music}}{\text{S}_{text}+\text{S}_{music}}
\end{align}

\section{Dance Design Studio}
\label{sec:studio}
For artists, a dedicated, lightweight choreography design studio is important because it converts the generative AI system from a "black box" into an active, collaborative, and visually pleasing tool. Choreography requires semantically meaningful control and an iterative creative process that must operate in the language of dance—techniques, gestures, and stylistic intentions—not raw data. By providing direct, interpretable, and fine-grained control, the studio empowers creators to direct the "what" and "why" of movement, supporting their workflow and ensuring the technology serves as a partner rather than a replacement.

\begin{figure}[t]
    \centering
    \includegraphics[width=\textwidth]{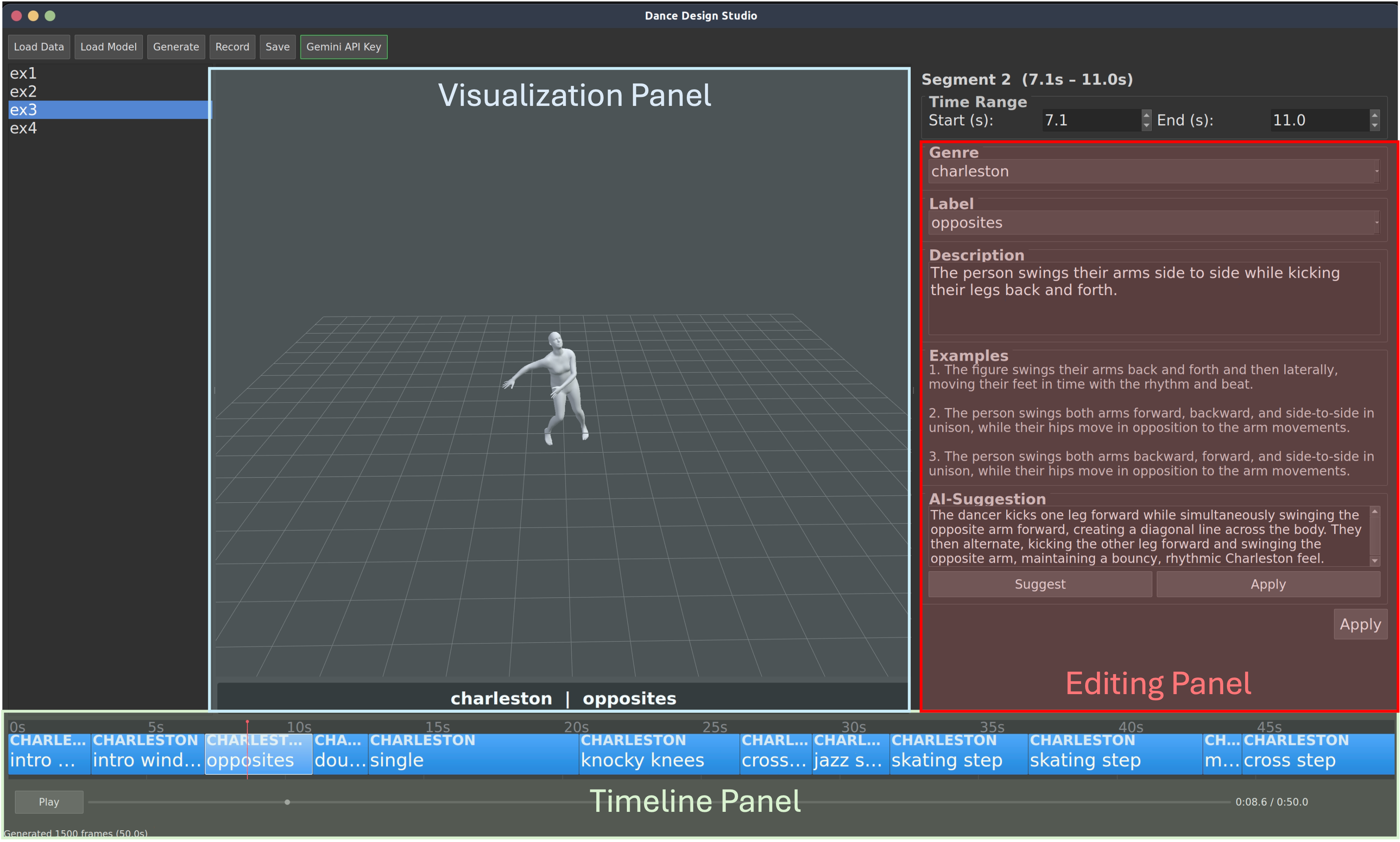}
    \caption{Collaborative Dance Design Studio: A Collaborate choreography environment Tool. This conceptual Dance Design Studio demo consists of three main panels: Visualization panel (blue), Timeline panel (green), and Editing panel (red). The Visualization panel displays a 3D generated character along with the current dance genre (\textit{Charleston}) and dance technique (\textit{Opposites}). The Timeline panel shows the performance timeline and how the user structures the current choreography. Finally, the Editing panel contains selectable Genre and Label lists with an editable text description. In addition, the system provides an AI-based suggestion feature that generates possible text descriptions. Using this tool, choregraphers can easily prototype their performance.}
    \label{fig:studio_concept}
\end{figure}
The studio follows a video editing software layout (see Fig.~\ref{fig:studio_concept}). It features a side bar that controls the actions of motion generation, and a timeline-like design at the bottom. Users can click on any moment to edit the label and the associated motion. The 3D canvas provides a powerful visual channel, allowing users to view the motion from any angle. Through the studio, dancers are able to swap the current music with a song of a different genre. If they are not satisfied with a specific label or the entire dance motion sequence, they can either ask the label generative model to produce a series of new motions or edit the existing one. Dancers can not only select labels from the existing collection of motion primitives, but they can also create or define their own motions.

Ultimately, the Dance Design Studio translates our Energy-based Diffusion Model's technical capabilities into a practical and artist-centric interface. By integrating music alignment, semantic text control, and flexible editing within this unified environment, we demonstrate a robust path toward controllable dance generation, positioning AI as a vital new instrument for choreographic creativity.

\end{document}